\documentclass[10pt,twocolumn,letterpaper]{article}

\usepackage[utf8]{inputenc}
\usepackage[T1]{fontenc}

\usepackage{iccv}
\usepackage{times}
\usepackage{epsfig}
\usepackage{graphicx}
\usepackage{amsmath}
\usepackage{amssymb}
\usepackage{multirow}

\usepackage[pagebackref=true,breaklinks=true,letterpaper=true,colorlinks,bookmarks=false]{hyperref}

\iccvfinalcopy 

\def\iccvPaperID{346} 

\usepackage[capitalise]{cleveref}
\Crefname{figure}{Figure}{Figures}
\Crefname{equation}{Equation}{Equations}
\usepackage{tikz,pgfplots}
\usetikzlibrary{backgrounds,calc,external}
\pgfplotsset{compat=1.4}

\usepackage{fancyhdr}
\newcommand{\G}{\mathcal{G}}

\newcommand{\DD}{\mathrm{D}}
\newcommand{\PP}{\mathrm{P}}
\newcommand{\FF}{\mathrm{F}}
\newcommand{\m}{\boldsymbol{\mu}}
\newcommand{\s}{{\sigma}}
\newcommand{\cam}{\mathbf{o}}
\newcommand{\x}{\mathbf{x}}
\newcommand{\n}{\mathbf{n}}
\newcommand{\an}{\mathbf{a}}
\newcommand{\T}{\boldsymbol{\gamma}}
\newcommand{\TT}{\boldsymbol{\theta}}
\newcommand{\I}{\mathbf{I}}
\newcommand{\IN}{\hspace{-0.1cm}\in\hspace{-0.1cm}}
\newcommand{\INN}{\hspace{-0.07cm}\in\hspace{-0.07cm}}
\DeclareMathOperator\erf{erf}
\newcommand*\dif{\mathop{}\!\mathrm{d}} 

\newcommand{\dt}{\dif t}
\newcommand{\ds}{\dif s}
\newcommand{\mR}{{\bar{\mu}}}
\newcommand{\sR}{{\bar{\sigma}}}
\newcommand{\cR}{{\bar{c}}}
\newcommand{\V}{\mathcal{V}}

\begin{document}

\title{A Versatile Scene Model with Differentiable Visibility\\Applied to Generative Pose Estimation}


\author{%
Helge Rhodin$^\text{1}$\hspace{0.8em}
Nadia Robertini$^\text{1, 2}$\hspace{0.8em}
Christian Richardt$^\text{1, 2}$\hspace{0.8em}
Hans-Peter Seidel$^\text{1}$\hspace{0.8em}
Christian Theobalt$^\text{1}$%
\\\vspace*{1pt}\\
$^\text{1}$ MPI Informatik\quad%
$^\text{2}$ Intel Visual Computing Institute%
}

\maketitle

\begin{abstract}
Generative reconstruction methods compute
 the 3D configuration (such as pose and/or geometry) of a shape by optimizing the overlap of the projected 3D shape model with images.
Proper handling of occlusions is a big challenge, since the visibility function that indicates if a surface point is seen from a camera can often not be formulated in closed form, and is in general discrete and non-differentiable at occlusion boundaries.
We present a new scene representation that enables an analytically differentiable closed-form formulation of surface visibility.
In contrast to previous methods, this yields smooth, analytically differentiable, and efficient to optimize pose similarity energies with rigorous occlusion handling,
fewer local minima,
and experimentally verified improved convergence of numerical optimization. 
The underlying idea is a new image formation model that represents opaque objects by a translucent medium with a smooth Gaussian density distribution which turns visibility into a smooth phenomenon.
We demonstrate the advantages of our versatile scene model in several generative pose estimation problems, namely marker-less 
multi-object pose estimation, marker-less human motion capture with few cameras, and image-based 3D geometry estimation.
\end{abstract}

\section{Introduction}

Many vision algorithms employ a generative approach to estimate the configuration $\TT$ of a 3D shape that optimizes a function measuring the 
similarity of the projected 3D model with one or more input camera views of a scene. 
In rigid object tracking, for example, $\TT$ models the global pose and orientation of an object, whereas in generative marker-less human motion capture, $\TT$ instead models the skeleton pose, and optionally surface geometry and appearance.

The ideal objective function for optimizing similarity has several desirable properties that are often difficult to satisfy: it should have analytic form, analytic derivative, exhibit few local minima, be efficient to evaluate, and numerically well-behaved, \ie smooth.
Many approaches already fail to satisfy the first condition and use similarity functions that cannot be expressed or differentiated analytically.
This necessitates the use of computationally expensive particle-based optimization methods or numerical gradient approximations that may cause 
instability and inaccuracy.

 
A major difficulty in achieving the above properties is the handling of occlusions when projecting from 3D to 2D.
Only those parts of a 3D model visible from a camera view should contribute to the similarity.
In general, this can be handled by using a visibility function $\V(\TT)$ in the similarity measure that describes the visibility of a surface point in pose $\TT$ when viewed from a certain direction.
For many shape representations, this function is unfortunately not only hard to formulate explicitly, but it is also binary for solid objects, and hence non-differentiable at points along occlusion boundaries.
This renders the similarity function non-differentiable.

In this paper, we introduce a 3D scene representation and image formation model 
that holistically addresses visibility within a generative similarity energy.
It is the first model that satisfies all the following properties: 
%
\begin{enumerate}
\item It enables an 
analytic, continuous and smooth visibility function that is differentiable everywhere in the scene.
\item It enables similarity energies with rigorous visibility handling that are differentiable everywhere in the model and camera parameters.
\item It enables similarity energies that can be optimized efficiently with gradient-based techniques, and which exhibit favorable and more robust convergence in cases where previous visibility approximations fail, such as disocclusions or multiple nearby occlusion boundaries.
\end{enumerate}
Our method approximates opaque objects by a translucent medium with a smooth density distribution defined via a collection of Gaussian functions. This turns occlusion and visibility into smooth phenomena.
Based on this representation, we derive a new rigorous image formation model that is inspired by the principles of light transport in translucent media common in volumetric rendering \cite{Cerezo2005},
and which ensures the advantageous properties above. 

Although visibility of solid objects is non-differentiable by nature, we demonstrate experimentally in \cref{sec:results} that introducing approximations on the scene level is advantageous compared to state-of-the-art methods that employ binary visibility and use spatial image smoothing.
We demonstrate the advantages of our approach in several scenarios: marker-less human motion capture with a low number of cameras compared to state-of-the-art methods that lack rigorous visibility modeling~\cite{Stoll2011,Elhayek2015}, body shape and appearance estimation from silhouettes, and
 more robust multi-object pose optimization compared to methods using local visibility approximations \cite{Loper2014}.



\section{Related work}


For numerical optimization of generative model-to-image similarity, the objective function needs to consider surface visibility, and needs to be differentiated.
The problem is that the (discrete) visibility function $\V$ is generally non-differentiable at occlusion boundaries of solid objects, and often hard to express in analytic form.
Some approaches avoid explicit construction of $\mathcal{V}$ by heuristically fixing occlusion relations at the beginning of iterative numerical optimization, which can easily lead to convergence to erroneous local optima, or
by re-computation of visibility before each iteration, which can become computationally prohibitive.
Commonly the object's silhouette boundaries are handled as special cases, different from the shape interior~\cite{Matusik2000,Starck2007,Tung2009,ChengleiICCV11}.
These approaches optimize the model configuration (\eg pose and/or shape) such that the projected model boundaries align with 
multi-view input silhouette boundaries (and possible additional features away from the silhouette), \eg~\cite{Deutscher2005,Rosenhahn2005,Urtasun:2006,Sigal:2010,InteractingHands12}.

\begin{figure}[t!]
\begin{center}
\begin{tikzpicture}[tight background,
  image/.style={inner sep=0pt},
  subcaption/.style={inner xsep=0.75mm, inner ysep=0.75mm, scale=0.6, above right},]
\def\padding{2pt}
\newcommand{\subfig}[2]{\includegraphics[width=0.2\linewidth,angle=90,origin=c,trim=30px 45px 30px 45px, clip]{images/movingSphere/#1_#2}}

\node[image]                (hard-f150) at (.06,8)          {\subfig{solidSphere}{f150}};
\node[image,right=\padding] (hard-f200) at (hard-f150.east) {\subfig{solidSphere}{f200}};
\node[image,right=\padding] (hard-f250) at (hard-f200.east) {\subfig{solidSphere}{f250}};
\node[image,right=\padding] (hard-f300) at (hard-f250.east) {\subfig{solidSphere}{f300}};
\node[image,right=\padding] (hard-f350) at (hard-f300.east) {\subfig{solidSphere}{f350}};

\draw[ultra thick,black!70] (hard-f150.center) ++(-2mm, 2.2mm) -- +(4mm,-4mm);
\draw[ultra thick,black!70] (hard-f150.center) ++(-2mm,-1.8mm) -- +(4mm, 4mm);

\foreach \img in {f150,f200,f250,f300,f350}
{
	\node[image,below=\padding] (smoothed-\img) at (hard-\img.south) {\subfig{solidSphere_gauss}{\img}};
	\node[image,below=\padding] (ours-\img) at (smoothed-\img.south) {\subfig{densitySphere_k01}{\img}};
}

\node[subcaption,white] at (hard-f150.south west)     {\textsf{Binary}};
\node[subcaption,white] at (smoothed-f150.south west) {\textsf{Smoothed}};
\node[subcaption,white] at (ours-f150.south west)     {\textsf{Our visibility}};

\node[subcaption,above,black] at (hard-f150.north)   {\large{$\TT = 0.2$}};
\node[subcaption,above,black] at (hard-f200.north)   {\large{$\TT = 0.4$}};
\node[subcaption,above,black] at (hard-f250.north)   {\large{$\TT = 0.6$}};
\node[subcaption,above,black] at (hard-f300.north)   {\large{$\TT = 0.8$}};
\node[subcaption,above,black] at (hard-f350.north)   {\large{$\TT = 1.0$}};

	\begin{axis}[footnotesize, width=\columnwidth, height=4.5cm,
		xmin=125, xmax=425, xlabel={Vertical sphere position $\TT$}, 
		xtick={100,150,...,500},
		xticklabels={$0$,$0.2$,$0.4$,$0.6$,$0.8$,$1$,$1.2$},
		ymin=0, ymax=1.1, ylabel={Visibility $\V(\TT)$}, ytick={0,0.2,...,1},
yticklabel style={/pgf/number format/fixed,  /pgf/number format/precision=2},
		legend columns=1, legend entries={binary, partial, smoothed, ours},
		legend style={text width=1cm,cells={anchor=west}},
		table/every table/.style={header=false,x expr=\coordindex,clip marker paths=true}
	]%

\coordinate (axis-f150) at (axis cs:150, 1.1);
\coordinate (axis-f200) at (axis cs:200, 1.1);
\coordinate (axis-f250) at (axis cs:250, 1.1);
\coordinate (axis-f300) at (axis cs:300, 1.1);
\coordinate (axis-f350) at (axis cs:350, 1.1);

\addplot[red!50,  mark=none, mark=square*, only marks] table[y index=1] {images/data/visibilityFunctions_k01.txt};%
\addplot[red!70!black,   mark=none, dashed] table[y index=3] {images/data/visibilityFunctions_k01.txt};%
\addplot[green!50!black, mark=none, semithick] table[y index=5] {images/data/visibilityFunctions_k01.txt};%
\addplot[blue!70!black,  mark=none, semithick] table[y index=7] {images/data/visibilityFunctions_k01.txt};%

\end{axis}
\foreach \x in {f150,f200,f250,f300,f350} \draw[black,thick] (axis-\x) -- (ours-\x.south);
\end{tikzpicture}
\end{center}\vspace{-.5em}
\caption{Visibility comparison on a vertically moving sphere.
Top to bottom: solid scene with binary visibility, spatial image smoothing, and our visibility model for positions $\TT \!\in\! \{0.2,\!0.4,\!0.6,\!0.8,\!1\}$.
Bottom: plot of the red sphere's visibility at the central pixel (marked by the gray cross in the first image) versus sphere position $\TT$ for the different visibilities.
Only our method is smooth at the double occlusion boundaries at $\TT = 0.4$ and $\TT=0.8$.}
\label{fig:smoothVisibilityAnalysis}
\end{figure}
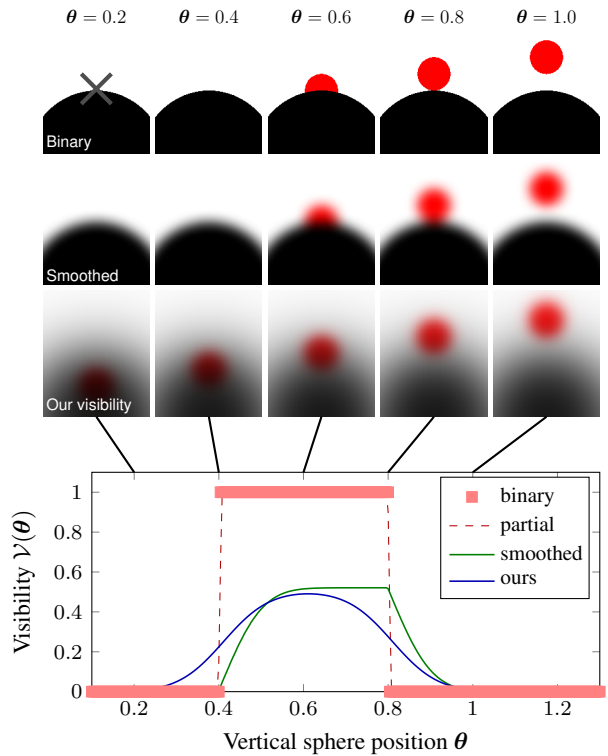

The integration of binary visibility into the similarity function is more complex.
Analytic visibility gradients can be obtained for implicit shapes \cite{Gargallo2007} and mesh surfaces \cite{Delaunoy2011}, by resorting to distributional derivatives \cite{Yezzi2003,Gargallo2007,Delaunoy2011}, and by geometric considerations on the replacement of a surface with another with respect to motions of the occlusion boundary \cite{Jalobeanu2004,DeLaGorce2008,Loper2014}.
For multi-view reconstruction of convex objects, visibility can be inferred efficiently from surface orientation \cite{Lempitsky2006}.
While these approaches yield similarity functions that are mathematically differentiable almost everywhere, non-differentiability is resolved only locally, which still leads to abrupt changes of object visibility, as illustrated in \cref{fig:smoothVisibilityAnalysis}.
Efficient gradient-based numerical optimization of the similarity does not fare well under such abrupt and localized changes, leading to frequent convergence to erroneous local optima.

\emph{OpenDR} uses binary visibility, provides  
an open-source renderer that models illumination and appearance of arbitrary mesh objects, and computes numerically approximated derivatives with respect to the model parameters for perspective projection~\cite{Loper2014}.
Finite differences are used for the spatial derivatives of pixel colors, as proposed for faces by Jones and Poggio \cite{Jones1996}.
To attain smooth visibility at single occlusion boundaries, spatial smoothing by convolution of the model projection with a smooth kernel \cite{Jones1996,Yezzi2003}, and coarse-to-fine pyramid representations \cite{Jones1996,Blanz1999,Loper2014} are used. 
Some global dependencies in pose energy between distant scene elements are also handled by coarse-to-fine approaches.
Our scene and visibility model handles such global effects by design and is the only model that handles the important case of double occlusion boundaries well, \eg at the point of complete occlusion of an object, see \cref{fig:smoothVisibilityAnalysis}.
Some recent methods abandon the use of surface representation and instead employ an implicit 3D shape model for performing tracking of full-body motion \cite{plankers_articulated_2003,Moeslund2006,Stoll2011,Elhayek2015}, hand motion \cite{Bretzner2002,Sridhar2015} and object poses \cite{Ren2014}. 
The implicit surfaces can be considered smooth at reprojection boundaries and are therefore well-suited for modeling a differentiable, well-behaved visibility function. 
Stoll \etal \cite{Stoll2011} do marker-less skeletal pose optimization from multi-view video, and use a collection of volumetric
3D Gaussians
to represent the human body, as well as 2D Gaussians to model the images.
A coarse occlusion heuristic thresholds the overlap between model and image Gaussians.
This design allows for long-range effects between model and observation, and avoids expensive occlusion tests, but leads to a problem formulation that is merely piecewise differentiable.
Follow-up work empowered tracking with a lower number of cameras by augmenting generative pose tracking with part detections in images~\cite{Elhayek2015}. 
None of these approaches uses a rigorous visibility model, as we do in this paper. 
\section{Scene model}

We propose a scene model that approximates solid objects by a smooth density representation, resulting in a visibility function that is well-behaved and differentiable everywhere.
In this section, we introduce our scene representation (§\ref{sec:sceneRepresenation}), give a physically-based intuition of the resulting visibility function in terms of a translucent medium (§\ref{sec:visibilityOfPoint}), and present the corresponding image formation model (§\ref{sec:imageFormation}).
Results of our scene model applied to rigid pose tracking and marker-less motion capture from sparse cameras are shown in \cref{sec:poseOptimization,sec:results}.




\subsection{Smooth scene approximation}
\label{sec:sceneRepresenation}

\begin{figure}
\begin{center}
\includegraphics[width=\linewidth]{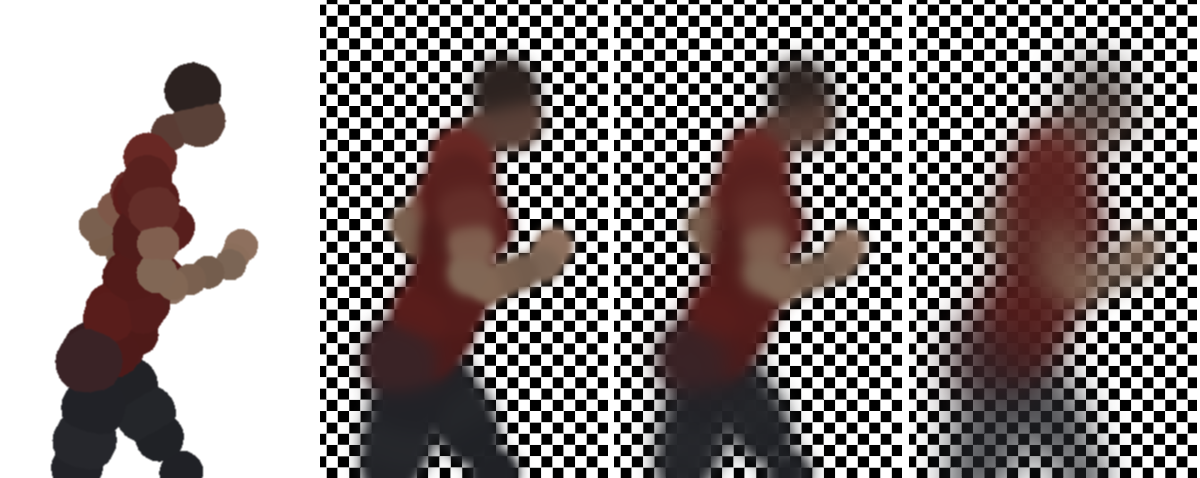}
\end{center}
   \caption{From left to right: A solid sphere actor model, our representation by a translucent medium with Gaussian density visualized on a checkerboard background for increasing smoothness levels ($m\!\!=\!\!\{0.0001,0.01,0.5\}$). Note the proper occlusion, \eg of the left arm and the torso.}
\label{fig:humanActorModel}
\end{figure}

Hard object boundaries cause discontinuities of visibility at occlusion boundaries. 
To obtain a smooth visibility function, we propose a smooth scene representation.
We diffuse objects to a smooth translucent medium – with high density at the inside of the original object and a smooth falloff to the outside. 
In our model, the density defines the extinction coefficient which models how opaque a point in space is, and thus how much it occludes \cite{Cerezo2005}.
To obtain an analytic form and for performance reasons, 
we use a parametric representation for the density $D(\x)$ at position $\x$ as the sum of scaled isotropic Gaussians $\G \!=\! \{ G_q\}_q$, defined as
\begin{align}
D(\x)
&= \sum_{G_q\in\G} G_q(\x), \text{ where each Gaussian} \nonumber \\
\label{eq:Gaussian}
G_q(\x) &=  c_q \cdot \exp\!\left(-\frac{ \left\| \x - \m_q \right\|^2}{2 \s_q^2}\right)
\end{align}
has a magnitude $c_q$, center $\m_q$ and standard deviation $\s_q$.
Appearance is modeled by annotating each Gaussian with an albedo attribute $\an_q$.
%
\Cref{fig:humanActorModel} shows an example of the colored density representation for a human actor, consisting of Gaussians of varying size and albedo.
%

Our model leads to a low-dimensional scene representation parametrized by $\T \!=\! \{c_q, \m_q, \s_q, \an_q \}_q$.
For readability, we use $G_q(\x)$ for Gaussians and omit the dependence on $\T$.

The degree of opaqueness and smoothness is adjustable by tuning the magnitudes $c_q$ and standard deviations $\s_q$ of the Gaussians.
We discuss the conversion of a general scene to our Gaussian density representation in \cref{sec:modelCreation}.
While other smooth basis functions are conceivable, Gaussians lead to
simple analytic expressions for the visibility that work well in practice.
While our Gaussian representation is similar to Stoll \etal's \cite{Stoll2011}, its semantics of a translucent medium is fundamentally different 
and our image formation model with rigorous visibility is phrased in entirely new ways, as explained in the following sections.

\subsection{Light transport and visibility}
\label{sec:visibilityOfPoint}

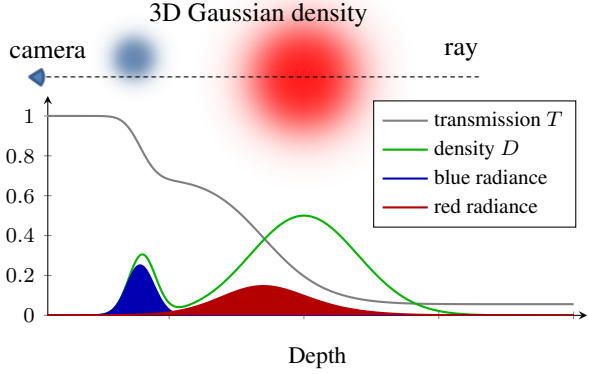
\begin{figure}\vspace{-0.5em}
\begin{center}
\begin{tikzpicture}[tight background]
\node at (2.85,3) {\includegraphics[width=0.9\linewidth,trim=0px 0px 0px 15px, clip]{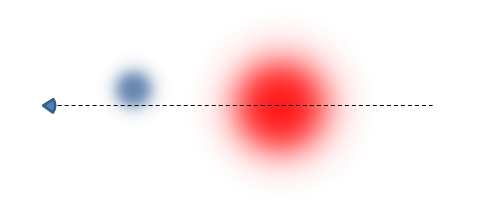}};
\node at (2.8,4) {3D Gaussian density};
\node at (0.0,3.5) {camera};
\node at (5.5,3.5) {ray};
	\begin{axis}[footnotesize, width=1.05\columnwidth, height=4.5cm,
		axis lines=left,
		xmin=20, xmax=820, 
		xlabel={Depth},
		xtick={0,200,...,800}, xtick pos=left,
		xticklabels=={,,}, 
		ymin=0, ymax=1.1, 
		ytick={0,0.2,...,1}, ytick pos=left,
		legend columns=1, legend entries={transmission $T$, density $D$, blue radiance, red radiance},
		legend style={cells={anchor=west}},
		table/every table/.style={header=false,x expr=\coordindex}
	]%
\addplot[black!50,  mark=none, thick] table[y index=0] {images/data/transmission.txt};%
\addplot[green!70!black,   mark=none, thick] table[y index=0] {images/data/density.txt};%
\addplot[blue!70!black, fill=blue!70!black, mark=none, thick] table[y index=0]  {images/data/radianceBlue.txt};%
\addplot[red!70!black, fill=red!70!black, mark=none, thick] table[y index=0] {images/data/radianceRed.txt};%
\end{axis}
\end{tikzpicture}
\end{center}\vspace{-.5em}
\caption{Top: Raytracing of a Gaussian density. Bottom: Light transport along the ray. 
The density along a ray is a sum of 1D Gaussians (green), and transmittance (gray) falls off from one for increasing optical depth. The radiance is the fraction of reflected light that reaches the camera (red and blue areas).
We use it to compute the visibility of a particular Gaussian.
}
\label{fig:densityTransmissionAndAbsorbtion}
\end{figure}

Our computation of the visibility $\V$ of a 3D point from a given camera position is inspired by the physical laws of light transport in translucent media, and based on simulation techniques from computer graphics \cite{Cerezo2005}.
As the translucent medium is only used as a tool to model continuous visibility, we assume a medium with uniform absorption of all colors without scattering.
According to the Beer-Lambert law of light attenuation, the transmittance (the percentage of light transmitted between two points in space) decays exponentially with the optical thickness of a medium, \ie the accumulated density, as visualized in \cref{fig:densityTransmissionAndAbsorbtion}. 
Specifically, the transmittance $T$ of a 3D point at distance $s$ along a ray from a camera position $\cam$ in direction $\n$ is 
\begin{align}
T(\cam,\n,s,\T) = \exp\!\left(- \int_0^s D(\cam+t \n) \dt \right) \text{.}
\label{eqn:transmittance_general}
\end{align}
Note that for a specific camera, $\n(u,v)$ is uniquely defined for each pixel location $(u,v)$;
from now on, we use the short notation $\n$ that is implicitly dependent on the pixel position.
With our Gaussian density representation, the density at any point on a line through a sum of 3D Gaussians is in turn the sum of 1D Gaussians.
Specifically, inserting the line equation $\x \!=\! \cam \!+\! s \n$ into the 3D Gaussian $G_q$ \labelcref{eq:Gaussian} results in a scaled 1D Gaussian of form $\cR\exp\!\left(-\frac{(x-\mR)^2}{2 \sR^2}\right)$, with $\mR = (\m-\cam)^\top \n$ and $\sR = \sigma$.
%
%
The updated magnitude is 
%
$\cR = c \cdot \exp\!\left(- \frac{(\m-\cam)^\top (\m-\cam) - \mR^2}{2 \sR^2}\right)$.
%
Using the Gaussian form of the density, we can rewrite the transmittance function \labelcref{eqn:transmittance_general}
in analytic form in terms of the error function, $\erf(s) = \frac{2}{\sqrt\pi}\int_0^s \exp(-t^2)\,\dt$, as
\begin{align}
T(\cam,\n, s, \T)
&= \exp\!\left(-\int_0^s \sum_q G_q(\cam+t \n) \dt \right) \label{eqn:GaussianTransmission} \\
&\hspace{-1.85cm}= \exp\!\left({\sum_q \frac{\sR_q \cR_q}{\sqrt{\frac 2 \pi}} \left(\erf\!\left(\!\frac{-\mR_q}{\sqrt{2} \sR_q}\!\right) \!-\! \erf\!\left(\!\frac{s \!-\! \mR_q}{\sqrt{2} \sR_q}\!\right) \right)}\right) 
\text{.}
\label{egn:visibility_point}
\end{align}
Similar formulations
are used for cloud rendering \cite{Zhou2007,Jakob2011}.
The transmittance of a medium is symmetric, it also measures the \emph{fractional visibility} of a point $\x \!=\! \cam \!+\! s \n$ from position $\cam$, which we denote by $\V(\x, \T) := T(\cam, \n, s, \T)$. 

\subsection{Image formation and Gaussian visibility}
\label{sec:imageFormation}


For image formation, we assume that all scene elements emit an equal amount of light $L_e$.
To produce a discrete image from the proposed Gaussian density model, we shoot a ray through each pixel of a virtual pinhole camera.
The pixel color is the fraction of source radiance that is emitted along the ray and reaches the camera (color and radiance 
are related by the camera transfer function; we assume a linear camera response and use pixel color and radiance interchangeably).
For the defined medium with pure absorption, the received radiance is the product of transmittance $T$, density $D$, albedo $\an$ and ambient radiance $L_e$, integrated along the ray $\x= \cam+s\n$,
\begin{align}
L(\cam,\n)
&= \! \int_{0}^{\infty} \!\! T(\cam,\n,s,\T) D(\x(s)) \an(\x(s)) L_e \ds.
\label{eqn:incomingRadiance}
\end{align}
This is a special form of the integrated \emph{radiative transfer equation}~\cite{Cerezo2005,Chandrasekhar1960}, and it models the fact that each point in space emits light proportional to its density $D(\x)$ and illumination $L_e$. 
For our Gaussian density with parameters $\T$ and fixed $L_e \!\!=\! 1$, we obtain
\begin{equation}
L(\cam,\n,\T)
= \int_{0}^{\infty} \!\!  T(\cam,\n,s,\T) \sum_q G_q(\cam+s\n) \an_q \ds.
\label{eqn:visibility_SAG_density}
\end{equation}
%
To obtain an analytic form, we approximate the infinite integral by sampling a compact interval $S_q \!=\! \{\mR_q \!+\! k \lambda_q \,|\, k \IN K \!\subset\! \mathbb{Z}\}$ around the mean of each $G_q$:
\begin{equation}
\hat{L}(\cam,\n, \T)
= \sum_q \an_q \sum_{s \in S_q} \lambda_q T(\cam,\n,s,\T)  G_q(\cam+s \n) 
\text{,}
\label{eqn:radiance_Sampled}
\end{equation}
where $\lambda_q \!\sim\! \sR_q$ is the sampling step length, which is adaptive to the Gaussian's size.

Gaussians have infinite support ($G_q(\x) \!>\! 0$ everywhere), but each Gaussian's contribution vanishes exponentially with the distance from its mean, so local sampling is a good approximation.
In practice, we found that five samples with $K \!=\! \{-4,-3,\ldots,0\}$ and $\lambda = \sR$ suffice.
Importance sampling could further enhance accuracy.

A final insight is that the inner sum in the radiance equation (\ref{eqn:radiance_Sampled}), the sum of the product of source radiance and transmission, measures the contribution of each Gaussian to the pixel color, and therefore computes the \emph{Gaussian visibility} 
\begin{align}
\V_q(\cam,\n,\T) :=  \sum_{s \in S_q}\lambda_q T(\cam,\n,s,\T)  G_q(\cam+s \n),
\label{eqn:GaussianVisibility}
\end{align}
of $G_q$ from camera $\cam$ in direction $\n$.
The Gaussian visibility from pixel $(u,v)$, $\V_q((u,v),\T)$, is 
equivalent to $\V_q(\cam,\n(u,v),\T)$.
The radiance $\hat{L}$ and Gaussian visibility $\V_q$ depend on the set of all Gaussians in the scene.
However, our model enables us to represent most scenes with a moderate number of Gaussians (§\ref{sec:modelCreation}), such that the analytic forms of $\hat{L}$ and $\V_q$ can be evaluated efficiently. For increased performance, we also exclude Gaussians with magnitude $\cR_q \!<\! 10^{-5}$ for the given ray direction, which does not impair tracking quality (see supplemental document).

\section{Model creation}
\label{sec:modelCreation}

In principle, arbitrary shapes can be approximated using a sufficiently large number of small, localized Gaussians.
We convert an existing mesh model to our representation by first filling the object's volume with spheres.
In our experiments, like the actor in \cref{fig:humanActorModel}, we place spheres manually, but automatic sphere packing could be used instead \cite{Wang2006}.

We then replace the spheres by Gaussians of `equal perceived extent'.
A translucent object forms its boundary at the point of strongest transmittance change.
To find suitable parameters $c_q$ and $\sigma_q$ that approximate a sphere of radius $r$, we place a Gaussian $G_q$ at its center and analyze the visibility $\V_q(\cam,\n,\{c_q,\sigma_q\})$ viewed from an orthographic camera (\ie $\n$ is fixed in view direction and $\cam$ is the pixel location).
We solve for magnitude $c_q$ and standard deviation $\sigma_q$ such that the transparency at the Gaussian's center, $1 - \V_q$, is equal to a constant $m$, and the inflection point of $\V_q$ lies at distance $r$ from the center.
Here, $m$ is a free parameter determining the level of smoothness and translucency, see \cref{fig:humanActorModel}.
This is a useful tool to tune robustness versus specificity, as we demonstrate in \cref{sec:modelComponents}.
This procedure aligns the perceived Gaussian size with the reference sphere outline while maintaining a consistent opacity across Gaussians of different size.
An example is \cref{fig:smoothVisibilityAnalysis}, where the inflection point is aligned with the binary occlusion boundary.
The optimization is necessary, as the inflection point of visibility deviates from the density's inflection point, and parameters $c_q$ and $\sigma_q$ jointly influence its location.

For generative tracking, the configuration of the tracked model $\TT$ needs to be mapped to our scene representation using a function $\T(\TT)$. 
In rigid object tracking, $\T(\TT)$ is a single rigid transform that determines the position $\m_q$ of all Gaussians $G_q$; the sizes $\s_q$ and densities $c_q$ are then fixed.
For skeletal motion capture, each Gaussian is rigidly attached to a bone in the skeleton, and $\TT$ represents global pose and joint angles. Other mappings 
for non-rigidly deforming shapes can be easily used, too.   

\section{Pose optimization}
\label{sec:poseOptimization}
Reconstruction methods based on our representation will compute $\TT$ from a set of image observations $\{\I_o\}_o$ captured from different camera positions $\cam$, by minimizing an objective function of the general form
\begin{align}
\FF(\TT,\{\I_o\}_o) = \sum_o \DD(\T(\TT),\I_o) + \PP(\TT),
\label{eqn:generalObjectiveFunction}
\end{align}
where $\DD(\T,\I)$ is a data term, \eg photo-consistency, and $\PP(\TT)$ is a prior on configurations, \eg a general regularization terms.
With our image formation model (§\ref{sec:imageFormation}), we can formulate a photo-consistency-based model-to-image overlap in a fully visibility-aware, yet analytic and analytically differentiable way as:
\begin{align}
\DD_\text{pc}(\T,\I) := \sum_{(u,v) \in I} \left\| \hat{L}(\cam, \n(u,v), \T) - \I(u,v) \right\|_2^2, 
\end{align}
%
where $\I(u,v)$ is the image color at pixel $(u,v)$. In \cref{sec:smoothVisibilityAnalysis}, we use the photo-consistency energy $\FF_\text{pc}(\TT,\{\I_o\}_o) = \sum_o \DD_\text{pc}(\T(\TT),\I_o)$ 
without prior for rigid object tracking and body shape and appearance estimation. 

We also demonstrate our approach for marker-less human motion capture. The generative method by 
Stoll \etal~\cite{Stoll2011} uses a Gaussian representation for the skeletal body model, and transforms the input image into 
a collection of Gaussians using color clustering. Their data term sums the color-weighted overlap of all image and projected model Gaussians using a scaled orthographic projection and without rigorous visibility handling, see their paper for details. 


For visibility-aware motion capture, we define a new pose energy $\FF_\text{mc}$ with a perspective camera model and a new visibility-aware data term that accumulates the color dissimilarity $d(\I(u,v),a_q)$ over all pixels $(u,v)$ in image $\I$ and Gaussians $G_q$, weighted by the Gaussian visibility $\V_q$: 
%
\begin{equation}
\DD_\text{mc}(\T,\I) \!=\!\! \sum_{(u,v)} \! \sum_q \! d(\I(u,v),a_q) \V_q(\cam,\n(u,v),\T) 
\text{.}
\label{eqn:dataTermBackprojection}
\end{equation}%
%
To analyze the influence of our new visibility function in isolation,
we adopt the remaining model components from the baseline method of Elhayek \etal \cite{Elhayek2014}.
To compensate for illumination changes, colors are represented in HSV space and the value channel is scaled by $0.2$.
To ensure temporal smoothness and anatomical joint limits, accelerations and joint limit violations are quadratically penalized in the prior term $\PP(\TT)$.
Motion capture with the new visibility-aware energy leads to significantly improved results, as we demonstrate in \cref{sec:performanceCaptureExperiments}. 

For all our experiments on rigid and articulated tracking, we utilize a conditioned conjugate gradient descent solver to minimize the objective function. 
The analytic derivatives of the objective functions $\FF_\text{pc}$ and $\FF_\text{mc}$ with respect to all parameters are listed in the supplemental document.



\section{Results}
\label{sec:results}

We first validate the advantageous properties of our model in general (§\ref{sec:smoothVisibilityAnalysis}), and then show how 
our scene representation and image formation model lead to improvements over the state of the art in rigid object tracking (§\ref{sec:objectTracking}), shape estimation, and marker-less human motion capture (§\ref{sec:performanceCaptureExperiments}).

\subsection{General validation}
\label{sec:smoothVisibilityAnalysis}

We validate the smoothness and global support of our visibility handling using a scene with simple occlusions: a red sphere, initially hidden by an occluder, moves up vertically and becomes visible (\cref{fig:smoothVisibilityAnalysis}).
In our model, the visibility $\V$ of the red sphere (a single Gaussian) is smooth, and hence differentiable with respect to position $\T$ (blue line).
This is in contrast to surface representations which are only piecewise differentiable: binary visibility functions have discontinuities (red line), and visibility with partial pixel coverage is continuous but non-differentiable at occlusion boundaries (dashed red line).
Methods that smooth pixel intensities spatially as a post-process obtain smoothness at single occlusion boundaries.
However, when an object occludes or disoccludes behind another occlusion boundary, like the red sphere becoming visible behind the black sphere, and thus two or more occlusion boundaries are in spatial vicinity, visibility is non-differentiable and localized (green line).
Improper handling of this case is a major limitation in practical applications, for instance in motion capture where an arm may disocclude from behind the body. 
Our approach handles these cases by considering near-visible objects, it `peeks' behind occlusion boundaries.

\subsection{Object tracking}
\label{sec:objectTracking}

\begin{figure}
\begin{center}
\begin{tabular}{@{}c@{ }c@{ }c@{ }c@{ }c@{ }c@{}}
\small Target & \small Initial & \multicolumn{2}{c}{\small Ours} & \multicolumn{2}{c}{\small OpenDR} \\
  &  pose & \small density &\small  mesh &\small  pyramid & \small per pixel\\
\includegraphics[width=0.15\linewidth]{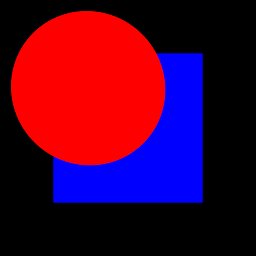} &
\includegraphics[width=0.15\linewidth]{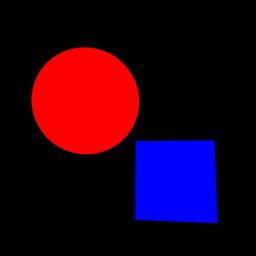} &
\includegraphics[width=0.15\linewidth]{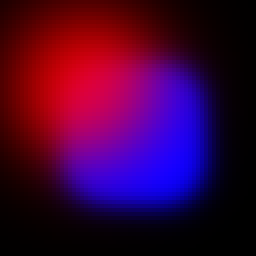} &
\includegraphics[width=0.15\linewidth]{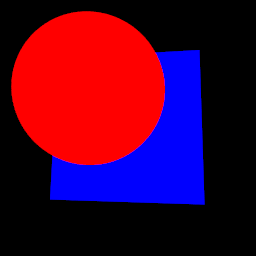} &
\includegraphics[width=0.15\linewidth]{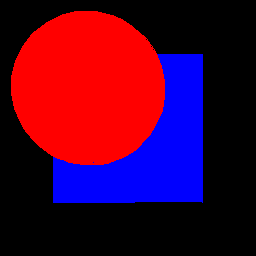}&
\includegraphics[width=0.15\linewidth]{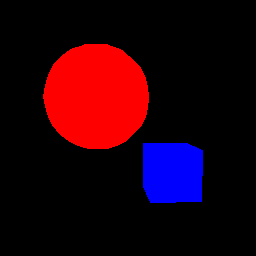}\\
 &
\includegraphics[width=0.15\linewidth]{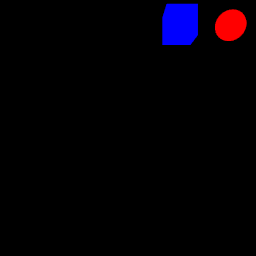} &
\includegraphics[width=0.15\linewidth]{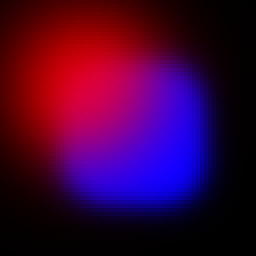} &
\includegraphics[width=0.15\linewidth]{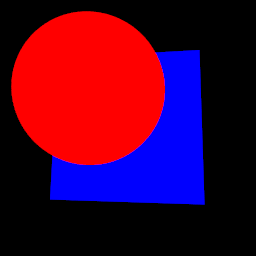} &
\includegraphics[width=0.15\linewidth]{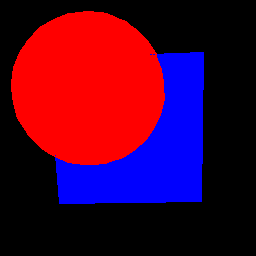}&
\includegraphics[width=0.15\linewidth]{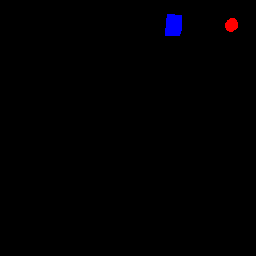}\\
 &
\includegraphics[width=0.15\linewidth]{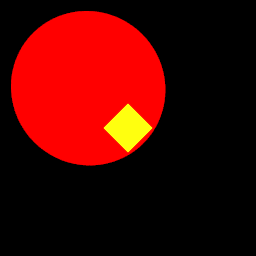} &
\includegraphics[width=0.15\linewidth]{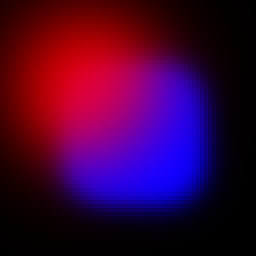} &
\includegraphics[width=0.15\linewidth]{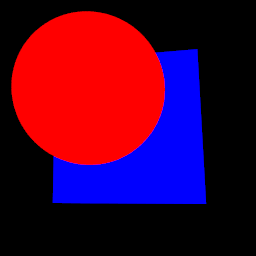} &
\includegraphics[width=0.15\linewidth]{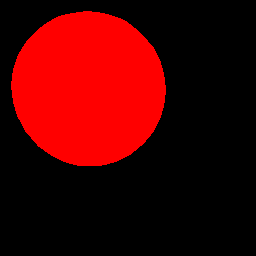} &
\includegraphics[width=0.15\linewidth]{images/OpenDR/OpenDR_fromOcclusion.png}\\
\end{tabular}
\end{center}
   \caption{3D reconstruction of a red sphere (position) and blue cube (position and orientation) using photo-consistency energy $\FF_\text{pc}$ from one image for three different initializations.
Top to bottom: initialization with overlap to final pose, distant initialization, occluded initialization (the occluded cube is shown in yellow).
OpenDR does not find the right solution for initializations without overlap or when far from the solution, and fails under full occlusions. Our method finds the correct pose in all three cases.}
\label{fig:comparisonOpenDR}
\end{figure}

We show the advantages of our representation for gradient-based multi-object pose optimization from a single view under photo-consistency and compare against OpenDR~\cite{Loper2014}.
The nine parameters in $\TT$ for the synthetic test scene are the 3D position of a red sphere, and position and orientation of the blue cube.
Both objects shall reach the pose shown in the \emph{Target} image of \cref{fig:comparisonOpenDR}.
We compare the optimization of $\FF_\text{pc}$ with $m \!=\! 0.1$ using our model, OpenDR with per-pixel photo-consistency, and OpenDR with a Gaussian pyramid of 6 levels.
The optimizer is initialized with 100 random and three manual cases (rows in \cref{fig:comparisonOpenDR}).
Without smoothing, OpenDR fails in all cases as object and observation boundary do not overlap sufficiently (last column).
With smoothing, OpenDR captures 70$\%$ of all random initializations, when one object is fully occluded it fails (fifth column, last row).
Our solution captures the correct pose in 88$\%$ of all random initializations, even under full occlusion if the occluded object is in the vicinity of the occlusion boundary (forth column, last row).
Only in few cases an erroneous local minimum is reached.
Averaged over all successful optimizations, the 3D Euclidean positional error of the sphere is $7\times10^{-3}$ times its diameter (ours) vs. $4\times10^{-5}$ (OpenDR) and for the cube $1.7\times10^{-2}$ (ours) vs.\ $1.3\times10^{-2}$ (OpenDR).
In essence, the density approximation (one Gaussian for the sphere, 27 Gaussians for the cube) increases robustness and is essential for certain scene configurations.
The inaccuracy due to the approximation of sharp edges is of small scale, $\approx 10^{-2}$ compared to OpenDR, model and observation align very well when visualized as meshes (fourth column).

We show in the supplemental document that our approach also enables accurate
generative geometry and appearance estimation of non-trivial 3D shapes from images.


\subsection{Marker-less human motion capture}
\label{sec:performanceCaptureExperiments}

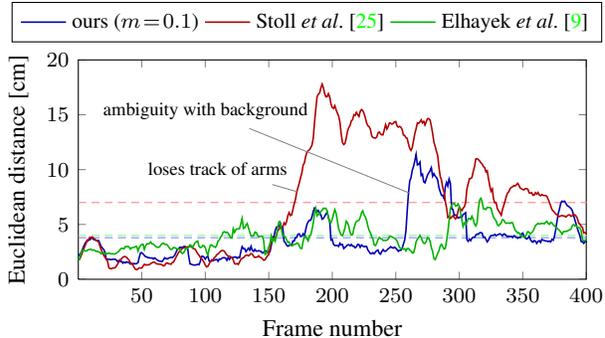
\begin{figure}
\begin{tikzpicture}[tight background]
	\begin{axis}[footnotesize, width=\columnwidth, height=4.5cm,
		xmin=0, xmax=400, xlabel={Frame number}, xtick={50,100,...,400},
		ymin=0, ymax=20, ylabel={Euclidean distance [cm]}, ytick={0,5,...,20},
		legend columns=3, legend entries={ours ($m\!=\!0.1$), Stoll \etal \cite{Stoll2011}, Elhayek \etal \cite{Elhayek2015}},
		legend style={anchor=south,at={(0.45,1.08)},cells={anchor=west}},
		table/every table/.style={header=false,x expr=\coordindex}
	]%
\addplot[blue!50,  mark=none, densely dashed, forget plot] table[y index=3] {images/data/ms_walk_relatedWorkComparison.txt};%
\addplot[red!50,   mark=none, densely dashed, forget plot] table[y index=4] {images/data/ms_walk_relatedWorkComparison.txt};%
\addplot[green!50, mark=none, densely dashed, forget plot] table[y index=5] {images/data/ms_walk_relatedWorkComparison.txt};%
\addplot[blue!70!black,  mark=none, semithick] table[y index=0] {images/data/ms_walk_relatedWorkComparison.txt};%
\addplot[red!70!black,   mark=none, semithick] table[y index=1] {images/data/ms_walk_relatedWorkComparison.txt};%
\addplot[green!70!black, mark=none, semithick] table[y index=2] {images/data/ms_walk_relatedWorkComparison.txt};	%
\node[coordinate,pin={[pin distance= 1mm]100:{\scriptsize loses track of arms}}] at (axis cs:173, 7.9619) {};%
\node[coordinate,pin={[pin distance=15mm]145:{\scriptsize ambiguity with background}}] at (axis cs:261, 7.8527) {};%
\end{axis}
\end{tikzpicture}
   \caption{Reconstruction accuracy against marker-based ground truth. Stoll \etal looses track of the arms after frame 150, and Elhayek \etal lacks accuracy during the first half of the sequence. Our method has a 3.7\;cm average joint position error – 45\% better than Stoll \etal with 7\;cm and best overall (dashed lines).}
\label{fig:markerBasedComparison}
\end{figure}

We now show the benefits of our approach for marker-less human motion capture on three multi-view video sequences with single and multiple actors\footnote{A table describing each scene and the relevant parameters, such as number, type and resolution of cameras, how many actors, pose parameters, run time etc., is given in the supplemental document.}.
Our approach optimizes $\FF_\text{mc}$ in the skeletal joint parameters (see \cref{sec:poseOptimization}).
We compare against the purely generative approach by Stoll \etal~\cite{Stoll2011}, and the recent combination of their  generative method with a ConvNet-based joint detection~\cite{Elhayek2015}, which was previously the only approach capable of marker-less skeletal motion capture in outdoor scenes with only 2–3 cameras. 

\begin{figure}
\begin{center}
\begin{tikzpicture}[tight background, image/.style={inner sep=0pt},]
\node[image]              (gen)  at (0,0)       {\includegraphics[width=0.24\linewidth]{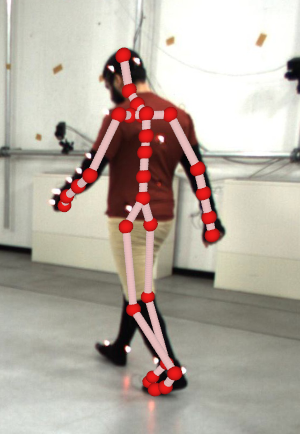}};
\node[image, right=0.5mm] (our1) at (gen.east)  {\includegraphics[width=0.24\linewidth]{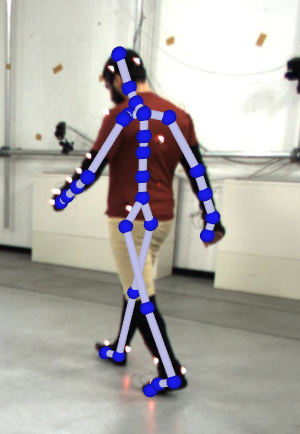}};
\node[image, right=2mm]   (disc) at (our1.east) {\includegraphics[width=0.24\linewidth]{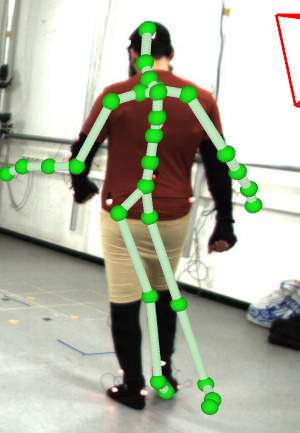}};
\node[image, right=0.5mm] (our2) at (disc.east) {\includegraphics[width=0.24\linewidth]{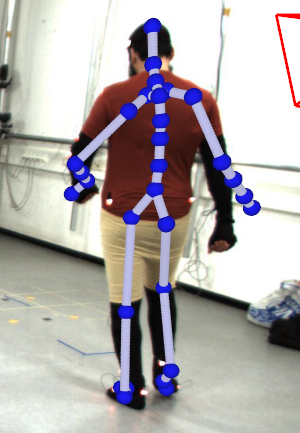}};

\node[below=-0.5mm] at (gen.south)  {\scriptsize Stoll \etal \cite{Stoll2011}};
\node[below=-0.5mm] at (our1.south) {\scriptsize ours ($m\!=\!0.1$)};
\node[below=-0.5mm] at (our2.south) {\scriptsize ours ($m\!=\!0.1$)};
\node[below=-0.5mm] at (disc.south) {\scriptsize Elhayek \etal \cite{Elhayek2015}};
\end{tikzpicture}
\end{center}\vspace{-.75em}
\caption{Pose estimates for the \emph{Marker} sequence, using two views for reconstruction.
Our method properly handles occlusion of the legs in frame 38 (left), and has much higher accuracy for frame 131 (right), here viewed from a third camera not used for tracking.}
\label{fig:markerBasedComparisonImages}
\end{figure}
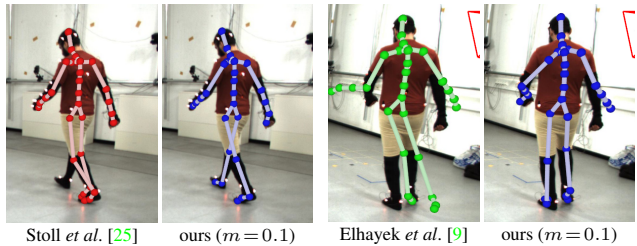

We first quantitatively compare against both methods using the average Euclidean reconstruction error against ground-truth 3D joint positions (from a concurrently run marker-based system) using two cameras of the indoor sequence 
\emph{Marker} \cite{Elhayek2015}, see \cref{fig:markerBasedComparison}.
All three algorithms use the same skeleton with 44 pose parameters, 72 Gaussians and data terms using HSV color space \cite{Elhayek2015} to be comparable. The implicit model 
needed for our approach is created as described in \cref{sec:modelCreation}.
Our method improves over the baseline \cite{Stoll2011} by a 45\% lower average error (3.7\;cm versus 7\;cm), as the baseline cannot track large parts of the sequence at all from only two views. 
Compared to Elhayek \etal, who use a discriminative component together with generative tracking, our method with visibility-aware, purely generative tracking is more precise for the first 250 frames and comparable for the last frames, where all three methods show errors due to ambiguities with the black background.
We thus achieve similarly robust marker-less captured with only two cameras as their more complex method. 
These quantitative improvements also manifest as clear qualitative pose improvements, as shown in \cref{fig:markerBasedComparisonImages}.
The proper visibility handling overcomes failures of previous techniques when arms and legs occlude.
Please see the supplemental material for videos.
We also show that the qualitative accuracy of our new marker-less approach captured with only four cameras on the \emph{Walker} sequence from \cite{Stoll2011} is comparable to their 12-camera result.

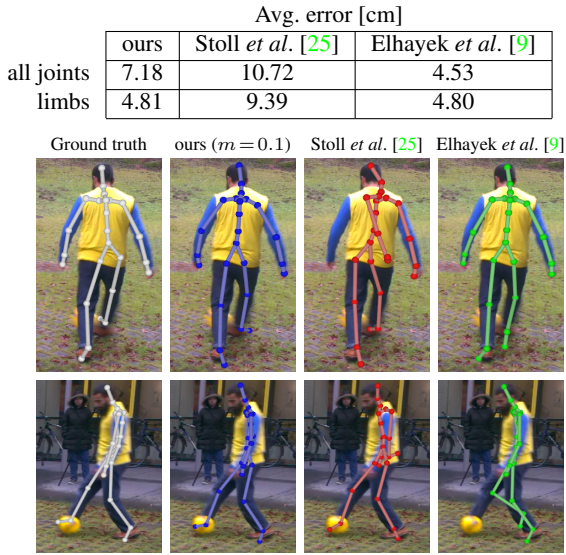
\begin{figure}
\begin{small}
\begin{tabular}{ r|c|c|c| }
\multicolumn{1}{r}{}
 &  \multicolumn{3}{c}{Avg. error [cm]}\\
 \cline{2-4}
 & ours &  Stoll \etal \cite{Stoll2011} & Elhayek \etal \cite{Elhayek2015} \\
\cline{2-4}
all joints & 7.18 & 10.72 & 4.53 \\
\cline{2-4}
limbs & 4.81 & 9.39 & 4.80 \\
\cline{2-4}
\end{tabular}
\end{small}
\vspace{-0.5em}
\begin{center}
\begin{tikzpicture}[tight background]

\newcommand{\subfigTop}[1]{\includegraphics[width=0.2\linewidth]{images/soccer/#1-crop}}
\newcommand{\subfigBottom}[1]{\includegraphics[width=0.2\linewidth]{images/soccer/#1-crop}}

\node[inner sep=0pt, anchor=south west] (gt) at (0,0) {\subfigTop{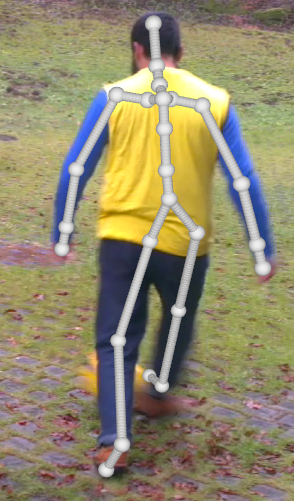}};
\node[inner sep=0pt, right=1mm] (our) at (gt.east) {\subfigTop{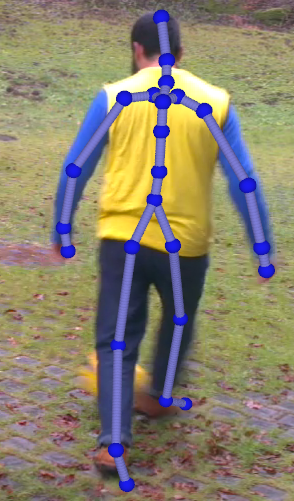}};
\node[inner sep=0pt, right=1mm] (gen) at (our.east) {\subfigTop{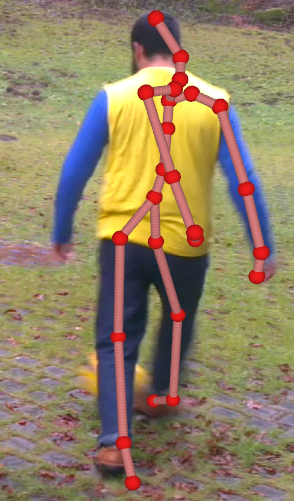}};
\node[inner sep=0pt, right=1mm] (comb) at (gen.east) {\subfigTop{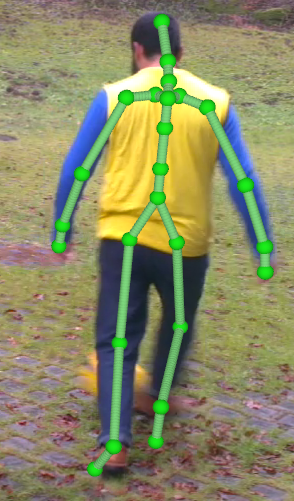}};

\node[inner sep=0pt, below=1mm] (gt1) at (gt.south) {\subfigBottom{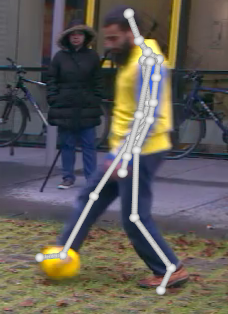}};
\node[inner sep=0pt, below=1mm] (our1) at (our.south) {\subfigBottom{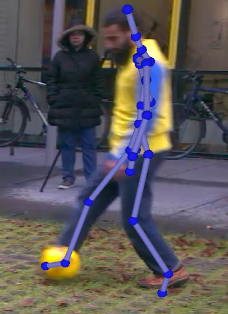}};
\node[inner sep=0pt, below=1mm] (gen1) at (gen.south) {\subfigBottom{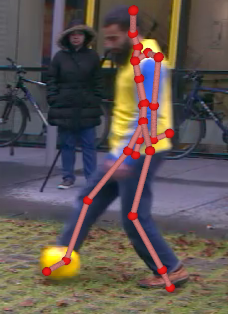}};
\node[inner sep=0pt, below=1mm] (comb1) at (comb.south) {\subfigBottom{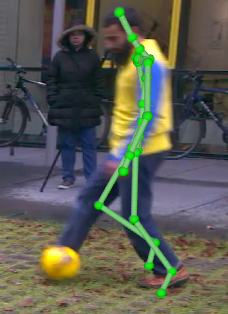}};

\node[above=1mm, anchor=base] at (gt.north)   {\scriptsize Ground truth};
\node[above=1mm, anchor=base] at (our.north)  {\scriptsize ours ($m\!=\!0.1$)};
\node[above=1mm, anchor=base] at (gen.north)  {\scriptsize Stoll  \etal \cite{Stoll2011}};
\node[above=1mm, anchor=base] at (comb.north) {\scriptsize Elhayek \etal \cite{Elhayek2015}};


\end{tikzpicture}
\end{center}

   \caption{Reconstruction accuracy for the \emph{Soccer} sequence against manually annotated ground truth and comparison to \cite{Stoll2011} and \cite{Elhayek2015}. The figure illustrates a case of ambiguity across two views, where the generative approach \cite{Stoll2011} looses track of the person's arm. Our approach and the approach from Elhayek \etal \cite{Elhayek2015} instead keep good average tracking with low 3D reprojection error for all joint positions, see the table above.
   }
\label{fig:SoccerSequence}
\end{figure}

We repeat the same comparison on the outdoor sequence \emph{Soccer}
with two actors, strong occlusions and fast actions, from only three views.
Again, we obtain significantly better accuracy than Stoll \etal in terms of 3D joint position, in particular for the limb joints (see \cref{fig:SoccerSequence}), as their results quickly show severe failures with so few cameras.
To analyze the impact of occlusions, we run our method once for a single actor, and in a second run we jointly track both actors (in total 84 parameters and 182 Gaussians). 
Simultaneous optimization not only handles self-occlusions but also the mutual occlusion of both actors.
This improves by 10.6\% and demonstrates the strength of precise and differentiable occlusion handling (see also \cref{fig:SoccerSequenceSingleAgainstTwoActorsVisual}).

In the supplemental document we also analyze the performance of our approach when evaluating our data term for cells of a quad tree that clusters pixels of similar color, as in Stoll \etal, instead of for all pixels of the input images .

\subsubsection{Radius of convergence}

\begin{figure}\centering
\begin{tikzpicture}[tight background]
	\begin{axis}[name=linePlot, footnotesize, width=0.7\columnwidth, height=5.5cm,
		xmin=-120, 
		xmax=25,
		 xlabel={\footnotesize Joint angle $\alpha$ [degrees]}, 
		xtick={-100,-75,...,100},
		ymin=0, ymax=1, ylabel={\footnotesize Energy $\FF_\text{mc}$}, ytick={0,0.2,...,1},
		legend columns=4, legend entries={distance,$m\!=\!0.5$,$m\!=\!0.1$,$m\!=\!0.0001$},
		legend style={anchor=south,at={(0.65,1.08)},cells={anchor=west}},
		table/every table/.style={header=false}
	]%
\addplot[black!50,  mark=none, densely dashed] table[x index={7}, y index={3}] {images/data/energyLandscape2D_shoulderJoint24.txt};%
\addplot[blue!70!black,  mark=none, semithick] table[y index=0, x index=7] {images/data/energyLandscape2D_shoulderJoint24.txt};%
\addplot[red!70!black,   mark=none, semithick] table[y index=1, x index=7] {images/data/energyLandscape2D_shoulderJoint24.txt};%
\addplot[green!70!black, mark=none, semithick] table[y index=2, x index=7] {images/data/energyLandscape2D_shoulderJoint24.txt};%
\end{axis}
\begin{axis}[name=barPlot, at={($(linePlot.east)+(0.3cm,0cm)$)}, anchor=west, footnotesize, width=0.3\columnwidth,height=5.5cm,
		ymin=50, ymax=80, enlargelimits=false, ytick={0,5,...,100},
 		xtick={-1},
 		ylabel={\footnotesize Successful convergence [\%]},
		x tick label style={rotate=50,anchor=east},
		ylabel near ticks, yticklabel pos=right
]%
\addplot [const plot,fill=green!70!black,draw=black] 
coordinates {(0,53)    (1,53) }  \closedcycle;%
\addplot [const plot,fill=red!70!black,draw=black] 
coordinates {(1,60)    (2,60) }  \closedcycle;%
\addplot [const plot,fill=blue!70!black,draw=black]
coordinates {(2,74)    (3,74) }  \closedcycle;%
\end{axis}
 \end{tikzpicture}
 \caption{A 1D slice through the energy landscape (for the shoulder joint angle) for different smoothness values $m$.
Higher values lead to a smoother energy with fewer local minima, and larger peaks further from the occlusion boundary (at $\alpha \!=\! -75$).
The global minimum of all configurations aligns well with the Euclidean distance to the ground truth (at $\alpha \!=\! -58$). Right: Convergences from 100 initializations within the shown interval.}
\label{fig:convergenceRadius}
\end{figure}
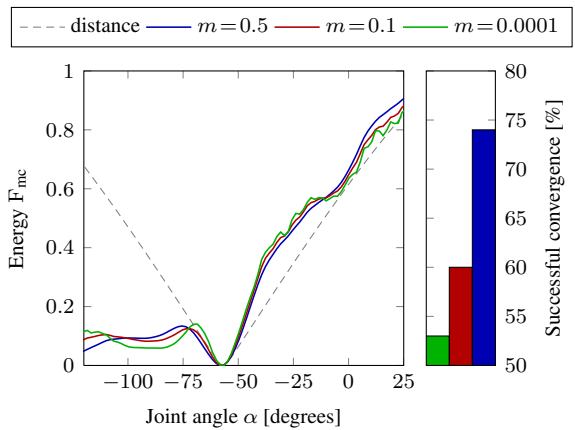

Our improved scene model with rigorous visibility handling leads to more \emph{well-behaved} similarity energies with a large \emph{radius of convergence}, \ie they converge for points further away from the global minimum, and a \emph{smooth energy landscape} with few local minima (already observed in OpenDR comparison).
We now validate these properties for the motion capture energy  
$\FF_\text{mc}$, see \cref{fig:convergenceRadius} left. 
For frame 83 of the \emph{Marker} sequence, we initialized the shoulder joint with $\alpha\IN[-127^{\circ}, 43^{\circ}]$ and analyzed different choices of $m$.
As expected, the energy $\FF_\text{mc}$ is smoother and contains fewer local minima for smaller values of $m$. 
We measure the convergence radius by optimizing from 100 initializations with $\alpha$ equally spaced over the shown interval and count successful convergences.
While all configurations succeed for initializations $\alpha\INN[-100^{\circ}, -20^{\circ}]$ close to the minimum $\alpha \!=\! -58^{\circ}$, only smoother versions ($m \!\geq\! 0.1$) converge for distant initializations, see \cref{fig:convergenceRadius} right.  The case $m \!=\! 0.0001$ models very sharp object boundaries, and hence gives results similar to methods with binary visibility. 

\begin{figure}

\begin{center}
\begin{tikzpicture}[tight background]
\node[anchor=south west] (gt1) at (0,0) {\includegraphics[width=0.4\linewidth]{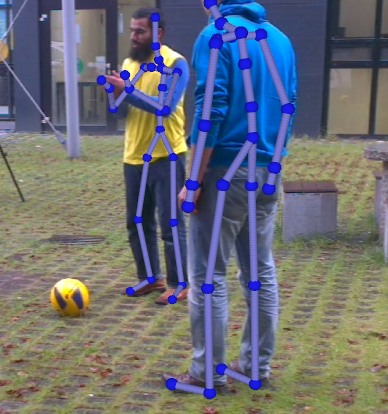}};
\node[anchor=south west] at (1,-0.3)  {\scriptsize ours ($2$ actors)};
\node[anchor=south west] (our1) at (3.7,0){\includegraphics[width=0.4\linewidth]{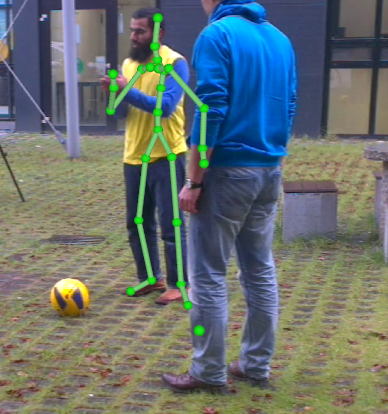}};
\node[anchor=south west] at (4.6,-0.3)        {\scriptsize ours ($1$ actor)};
\end{tikzpicture}
\end{center}\vspace{-.5em}
   \caption{Reconstruction for the {\it Soccer} sequence with comparison to tracking and modeling all versus a single actor. As shown in the figure, tracking of both subjects at the same time is advantageous as occluding regions are effectively handled by our approach.}
   \label{fig:SoccerSequenceSingleAgainstTwoActorsVisual}
\end{figure}

\subsubsection{Visibility gradient and smoothness level}
\label{sec:modelComponents}

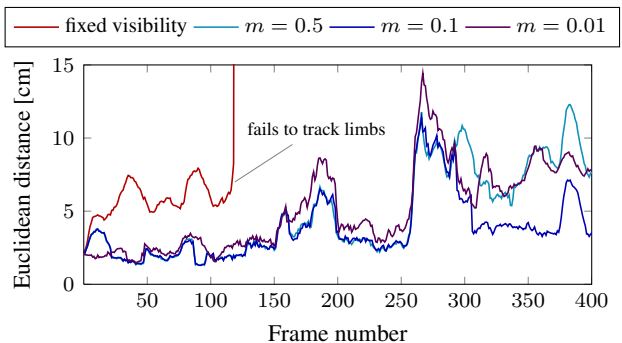
\begin{figure}
\begin{tikzpicture}[tight background]
	\begin{axis}[footnotesize, width=\columnwidth, height=4.5cm,
		xmin=0, xmax=400, xlabel={Frame number}, xtick={50,100,...,400},
		ymin=0, ymax=15, ylabel={Euclidean distance [cm]}, ytick={0,5,...,20},
		legend columns=4, legend entries={fixed visibility, $m=0.5$, $m=0.1$, $m=0.01$,},
		legend style={anchor=south,at={(0.45,1.08)},cells={anchor=west}},
		table/every table/.style={header=false,x expr=\coordindex}
	]%
\addplot[red!70!black,    mark=none, semithick] table[y index=1] {images/data/ms_walk_model.txt};
\addplot[cyan!70!black,   mark=none, semithick] table[y index=4] {images/data/ms_walk_model.txt};
\addplot[blue!70!black,   mark=none, semithick] table[y index=0] {images/data/ms_walk_model.txt};
\addplot[violet!70!black, mark=none, semithick] table[y index=3] {images/data/ms_walk_model.txt};
\node[coordinate,pin={[pin distance=5mm]85:{\scriptsize fails to track limbs}}] at (axis cs:119,    7) {};%
\end{axis}
\end{tikzpicture}
   \caption{Reconstruction accuracy of joints for different smoothness levels $m$, and for pre-computed fixed visibility per Gaussian.}
\label{fig:modelComponents}
\end{figure}

In our final experiment, we show that our new 
differentiable and well-behaved visibility function is essential for the success of our approach in marker-less human motion capture with very few cameras. 
For the \emph{Marker} sequence, we fix the visibility for each Gaussian and each camera prior to each iteration of the gradient-based optimizer,
\ie changes in occlusion are ignored during optimization. This setup quickly looses track of the limbs and fails completely after 110 frames, see \cref{fig:modelComponents}.
Moreover, to analyze the behavior of our method for different degrees of smoothness in our scene model, we compare multiple fixed smoothness levels.
The best trade-off between smoothness and specificity is attained for $m=0.1$. Which we use for all our experiments unless otherwise specified.

\subsubsection{Computational complexity and efficiency}

{
Our implementation of functions $\FF_\text{mc}$ and $\FF_\text{pc}$ and their gradients has complexity $O(N_I N_q^2 N_K \!+\! N_{\TT} N_q)$ for $N_I$ input pixels (summed over all views), and a scene of $N_q$ Gaussians, $N_K$ radiance samples and $N_{\TT}$ parameters.}
The quadratic complexity in terms of the number of Gaussians originates from the handling of multiple occlusion boundaries (occlusion test for each pair of Gaussians).
Our energy is nevertheless efficient to evaluate as the Gaussian density allows to model even complex objects, such as a human, by few primitives. For higher accuracy, coarse-to-fine approaches could be applied.
{
For the \emph{Marker} sequence the performance is 8.1 gradient iterations per second for $\FF_\text{mc}$, 10K input pixels (pixels far from the model do not contribute and are excluded), 72 Gaussians, and 44 pose parameters.
The experiments are executed for 200 iterations on a quad-core CPU with 3.6\;GHz. 
As the visibility evaluation of each pixel is independent, further speedups could be obtained by stochastic optimization and parallel execution on GPUs.}

\subsection{Discussion and conclusion}

We presented a new scene model and corresponding image formation model that approximates a scene by a translucent medium defined by Gaussian basis functions. This intentionally smoothes out shape and appearance. 
While this may introduce some uncertainty of shape models, it enables a visibility function and an image formation model that are differentiable everywhere, and efficient to evaluate. 
Analytic pose optimization energies were already used for motion capture \cite{Stoll2011,Elhayek2015}, but visibility was only approximated. 
Our new approach advances the state of the art by enabling analytic, smooth and differentiable pose energies with analytic and differentiable visibility. 
It also leads to larger convergence radii of these similarity energies.
This not only enables us to perform purely generative motion capture at the same accuracy but with far fewer cameras than Stoll \etal \cite{Stoll2011}, but also to achieve comparable accuracy with only 2--3 cameras as the more complex method by Elhayek \etal \cite{Elhayek2015} which combines generative and discriminative approaches.

OpenDR and other surface models may more accurately represent shape and texture, and also integrate light sources into the scene model.
This allows for higher alignment precision for some shapes, but it comes at the cost of a smaller convergence radius, failure under full occlusion, and lower computational efficiency than with our model. 

\section*{Acknowledgements}

We thank Oliver Klehm, Tobias Ritschel, Carsten Stoll and Nils Hasler for valuable discussions and feedback.
This research was funded by the ERC Starting Grant project CapReal (335545).

{\small
\bibliographystyle{ieee}
\bibliography{differentiableOcclusion}

\begin{thebibliography}{10}\itemsep=-1pt

\bibitem{InteractingHands12}
L.~Ballan, A.~Taneja, J.~Gall, L.~V. Gool, and M.~Pollefeys.
\newblock Motion capture of hands in action using discriminative salient
  points.
\newblock In {\em ECCV}, 2012.

\bibitem{Blanz1999}
V.~Blanz and T.~Vetter.
\newblock A morphable model for the synthesis of {3D} faces.
\newblock In {\em SIGGRAPH}, pages 187--194, 1999.

\bibitem{Bretzner2002}
L.~Bretzner, I.~Laptev, and T.~Lindeberg.
\newblock Hand gesture recognition using multi-scale colour features,
  hierarchical models and particle filtering.
\newblock In {\em Automatic Face and Gesture Recognition}, pages 423--428,
  2002.

\bibitem{Cerezo2005}
E.~Cerezo, F.~P\'{e}rez, X.~Pueyo, F.~J. Seron, and F.~X. Sillion.
\newblock A survey on participating media rendering techniques.
\newblock {\em The Visual Computer}, 21(5):303--328, 2005.

\bibitem{Chandrasekhar1960}
S.~Chandrasekhar.
\newblock {\em {Radiative Transfer}}.
\newblock Dover Publications Inc, 1960.

\bibitem{DeLaGorce2008}
M.~de~La~Gorce, N.~Paragios, and D.~J. Fleet.
\newblock Model-based hand tracking with texture, shading and self-occlusions.
\newblock In {\em CVPR}, 2008.

\bibitem{Delaunoy2011}
A.~Delaunoy and E.~Prados.
\newblock {Gradient flows for optimizing triangular mesh-based surfaces:
  Applications to 3D reconstruction problems dealing with visibility}.
\newblock {\em IJCV}, 95(2):100--123, 2011.

\bibitem{Deutscher2005}
J.~Deutscher and I.~Reid.
\newblock Articulated body motion capture by stochastic search.
\newblock {\em IJCV}, 61(2):185--205, 2005.

\bibitem{Elhayek2015}
A.~Elhayek, E.~Aguiar, A.~Jain, J.~Tompson, L.~Pishchulin, M.~Andriluka,
  C.~Bregler, B.~Schiele, and C.~Theobalt.
\newblock Efficient {ConvNet}-based marker-less motion capture in general
  scenes with a low number of cameras.
\newblock In {\em CVPR}, 2015.

\bibitem{Elhayek2014}
A.~Elhayek, C.~Stoll, K.~I. Kim, and C.~Theobalt.
\newblock Outdoor human motion capture by simultaneous optimization of pose and
  camera parameters.
\newblock {\em Computer Graphics Forum}, 2014.

\bibitem{Gargallo2007}
P.~Gargallo, E.~Prados, and P.~Sturm.
\newblock Minimizing the reprojection error in surface reconstruction from
  images.
\newblock In {\em ICCV}, 2007.

\bibitem{Jakob2011}
W.~Jakob, C.~Regg, and W.~Jarosz.
\newblock Progressive expectation--maximization for hierarchical volumetric
  photon mapping.
\newblock {\em Compututer Graphics Forum}, 30(4):1287--1297, 2011.

\bibitem{Jalobeanu2004}
A.~Jalobeanu, F.~O. Kuehnel, and J.~C. Stutz.
\newblock Modeling images of natural {3D} surfaces: Overview and potential
  applications.
\newblock In {\em CVPR Workshops}, pages 188--188, 2004.

\bibitem{Jones1996}
M.~J. Jones and T.~Poggio.
\newblock Model-based matching by linear combinations of prototypes.
\newblock A.I. Memo 1583, MIT, 1996.

\bibitem{Lempitsky2006}
V.~Lempitsky, Y.~Boykov, and D.~Ivanov.
\newblock Oriented visibility for multiview reconstruction.
\newblock In {\em ECCV}, 2006.

\bibitem{Loper2014}
M.~Loper and M.~J. Black.
\newblock {OpenDR}: An approximate differentiable renderer.
\newblock In {\em ECCV}, 2014.

\bibitem{Matusik2000}
W.~Matusik, C.~Buehler, R.~Raskar, S.~J. Gortler, and L.~McMillan.
\newblock Image-based visual hulls.
\newblock In {\em SIGGRAPH}, pages 369--374, 2000.

\bibitem{Moeslund2006}
T.~B. Moeslund, A.~Hilton, and V.~Kr\"{u}ger.
\newblock A survey of advances in vision-based human motion capture and
  analysis.
\newblock {\em CVIU}, 104(2):90--126, 2006.

\bibitem{plankers_articulated_2003}
R.~Plankers and P.~Fua.
\newblock Articulated soft objects for multiview shape and motion capture.
\newblock {\em PAMI}, 25(9):1182--1187, 2003.

\bibitem{Ren2014}
C.~Y. Ren, V.~Prisacariu, O.~Kaehler, I.~Reid, and D.~Murray.
\newblock {3D} tracking of multiple objects with identical appearance using
  {RGB-D} input.
\newblock In {\em 3DV}, pages 47--54, 2014.

\bibitem{Rosenhahn2005}
B.~Rosenhahn, C.~Perwass, and G.~Sommer.
\newblock Pose estimation of {3D} free-form contours.
\newblock {\em IJCV}, 62(3):267--289, 2005.

\bibitem{Sigal:2010}
L.~Sigal, A.~O. Balan, and M.~J. Black.
\newblock {HumanEva}: Synchronized video and motion capture dataset and
  baseline algorithm for evaluation of articulated human motion.
\newblock {\em IJCV}, 87(1--2):4--27, 2010.

\bibitem{Sridhar2015}
S.~Sridhar, F.~Mueller, A.~Oulasvirta, and C.~Theobalt.
\newblock Fast and robust hand tracking using detection-guided optimization.
\newblock In {\em CVPR}, 2015.

\bibitem{Starck2007}
J.~Starck and A.~Hilton.
\newblock Surface capture for performance based animation.
\newblock {\em IEEE Computer Graphics and Applications}, 27(3):21--31, 2007.

\bibitem{Stoll2011}
C.~Stoll, N.~Hasler, J.~Gall, H.-P. Seidel, and C.~Theobalt.
\newblock Fast articulated motion tracking using a sums of {Gaussians} body
  model.
\newblock In {\em ICCV}, pages 951--958, 2011.

\bibitem{Tung2009}
T.~Tung, S.~Nobuhara, and T.~Matsuyama.
\newblock Complete multi-view reconstruction of dynamic scenes from
  probabilistic fusion of narrow and wide baseline stereo.
\newblock In {\em ICCV}, pages 1709--1716, 2009.

\bibitem{Urtasun:2006}
R.~Urtasun, D.~J. Fleet, and P.~Fua.
\newblock {3D} people tracking with {Gaussian} process dynamical models.
\newblock In {\em CVPR}, pages 238--245, 2006.

\bibitem{Wang2006}
R.~Wang, K.~Zhou, J.~Snyder, X.~Liu, H.~Bao, Q.~Peng, and B.~Guo.
\newblock Variational sphere set approximation for solid objects.
\newblock {\em The Visual Computer}, 22(9--11):612--621, 2006.

\bibitem{ChengleiICCV11}
C.~Wu, K.~Varanasi, Y.~Liu, H.-P. Seidel, and C.~Theobalt.
\newblock Shading-based dynamic shape refinement from multi-view video under
  general illumination.
\newblock In {\em ICCV}, 2011.

\bibitem{Yezzi2003}
A.~Yezzi and S.~Soatto.
\newblock Stereoscopic segmentation.
\newblock {\em IJCV}, 53(1):31--43, 2003.

\bibitem{Zhou2007}
K.~Zhou, Q.~Hou, M.~Gong, J.~Snyder, B.~Guo, and H.-Y. Shum.
\newblock Fogshop: Real-time design and rendering of inhomogeneous,
  single-scattering media.
\newblock In {\em Pacific Graphics}, pages 116--125, 2007.

\end{thebibliography}
}

\end{document}


\title{Supplemental Document:\\
A Versatile Scene Model with Differentiable Visibility\\Applied to Generative Pose Estimation	}

\author{%
Helge Rhodin$^\text{1}$\hspace{0.8em}
Nadia Robertini$^\text{1, 2}$\hspace{0.8em}
Christian Richardt$^\text{1, 2}$\hspace{0.8em}
Hans-Peter Seidel$^\text{1}$\hspace{0.8em}
Christian Theobalt$^\text{1}$%
\\\vspace*{1pt}\\
$^\text{1}$ MPI Informatik\quad%
$^\text{2}$ Intel Visual Computing Institute%
}

\maketitle

\section{Introduction}

In our paper, we introduce a new visibility and image formation model that is differentiable everywhere. Applied to generative pose estimation, it improves convergence of numerical optimization. In this supplemental document, we explain in
more detail, and generality, the following parts of our main paper:

\begin{itemize}
\item Full details on the scene and model parameters of all our experiments (Section \ref{sec:results}).
\item Application of the visibility model to geometry and appearance estimation (Section \ref{sec:shapeOptimization}).
\item The qualitative comparison to Stoll \etal on the \emph{Walker} sequence \cite{Stoll2011} (Section \ref{sec:trackingQuality}).
\item Impact of varying the image resolution (Section \ref{sec:inputImageResolutin}).
\item Analytic gradients of visibility and introduced objective functions (Section \ref{sec:analyticGradients}).
\end{itemize}


\section{Results}
\label{sec:results}

\begin{figure}%
\begin{center}
\includegraphics[width=0.49\linewidth]{images/shapeMatching/ShapeReconstruction_images.jpg}
\includegraphics[width=0.49\linewidth]{images/shapeMatching/ShapeReconstruction_silhouettes}\\
\includegraphics[width=0.19\linewidth]{images/shapeMatching/image_c1_ff0000-crop}
\includegraphics[width=0.19\linewidth]{images/shapeMatching/image_c1_ff0100-crop}
\includegraphics[width=0.19\linewidth]{images/shapeMatching/image_c1_ff0300-crop}
\includegraphics[width=0.19\linewidth]{images/shapeMatching/image_c1_ff0637-crop}
\includegraphics[width=0.19\linewidth]{images/shapeMatching/image_c1_f2_i0-crop}
\end{center}
\caption{Shape and appearance estimation using the photo-consistency $\FF_\text{pc}$. Top: Input RGB images and extracted silhouettes. Bottom: reconstruction process from an unused camera view. From left to right: initialization, after 100, 300, 10000 iterations, and color back-projection. Each Gaussian is represented by a sphere of radius equal to its standard deviation.}
\label{fig:shapeReconstruction}
\end{figure}

Details on all input sequences, complexity of the utilized models, and optimization results are given in Table \ref{tbl:sequenceDetails}.
\begin{table*}[t]
  \centering
\begin{small}
\begin{tabular}{ |l|c|c|c|c|c|c|c|c| }
\hline
Sequence & 	                 \multicolumn{2}{|c|}{\emph{Soccer} (two actors)}	& \multicolumn{2}{|c|}{\emph{Soccer} (one actor)} & \emph{Marker} & \emph{Walker} & Shape estimate & Rigid objects\\ \hline\hline
Published by & 	\multicolumn{5}{|c|}{Elhayek \etal \cite{Elhayek2015}} & \multicolumn{2}{|c|}{Stoll \etal \cite{Stoll2011}} & Our\\ \hline
Number of cameras & 	\multicolumn{4}{|c|}{3} & 2 & 4 &11 & 1\\ \hline
Number of frames & 	\multicolumn{4}{|c|}{300} & 500 & 522 & \multicolumn{2}{|c|}{1}\\ \hline
Frame rate & 		\multicolumn{4}{|c|}{23.8} & 25 & 25 & \multicolumn{2}{|c|}{n/a}\\ \hline
Camera type & 		\multicolumn{4}{|c|}{mobile phone \emph{(HTC One X)}} & \multicolumn{3}{|c|}{PhaseSpace Vision Camera} & simulated\\ \hline
Raw image resolution & 	\multicolumn{4}{|c|}{1280$\times$720} & \multicolumn{3}{|c|}{1296$\times$972} & 256$\times$256 \\ \hline
Environment & \multicolumn{4}{|c|}{outdoor, uncontrolled background and lighting} & \multicolumn{3}{|c|}{studio, uncontrolled background} & synthetic\\ \hline
Tracked subjects &		\multicolumn{2}{|c|}{2}	& \multicolumn{2}{|c|}{1} & \multicolumn{3}{|c|}{1} & 2\\ \hline
Number of joints & 		\multicolumn{2}{|c|}{118}	 & \multicolumn{2}{|c|}{59} & 61 & 66 & \multicolumn{2}{|c|}{0}\\ \hline
Number of parameters  &	\multicolumn{2}{|c|}{84}	& \multicolumn{2}{|c|}{42} & 44 & 43 & 800 & 9\\ \hline
Number of Gaussians  & 		\multicolumn{2}{|c|}{182}	& \multicolumn{2}{|c|}{72} & 72 & 77 & 200 & 28\\ \hline
Input image resolution & 	320$\times$180 & 640$\times$360 & 320$\times$360 & 640$\times$360 & \multicolumn{3}{|c|}{324$\times$243} & 128$\times$128\\ \hline
Input pixels per frame & 	$\approx$12,000 & $\approx$202,000 & $\approx$7,000 & $\approx$122,000 & $\approx$10,000  & $\approx$25,000 & $\approx$80,000 & $\approx$16,000\\ \hline
Ground truth & \multicolumn{4}{|c|}{Manual annotation, 3D triangulation} & Marker & 12 cam & n/a & constructed \\ \hline
Average error [cm] & 4.81 & 4.69 & 5.88 & 4.70 & 3.79  & 2.55 & \multicolumn{2}{|c|}{n/a} \\ \hline
Timing [iterations/s] & 3.33 & 0.23 & 10.0 & 0.86 & 8.1 & 5.01 & 0.68 &  2.14 \\ \hline
\end{tabular}
\end{small}
\caption{A table describing each scene and the relevant parameters, such as number, type and resolution of cameras, pose parameter, run time per gradient iteration, and reconstruction error (average Euclidean 3D distance over all joints and frames to the ground truth).}
\label{tbl:sequenceDetails}
\end{table*}

\subsection{Shape optimization}
\label{sec:shapeOptimization}

The goal of this experiment is to evaluate the applicability of our image formation model to geometry and appearance estimation from multiple RGB images.
Input to the method are 11 RGB images from calibrated cameras and corresponding silhouettes obtained by background subtraction (see Figure \ref{fig:shapeReconstruction}).
We initialize the model with 200 small white Gaussians positioned randomly around the center of the capture volume. Subsequently, the position $\m_q$ and size $\sigma_q$ of each Gaussian $G_q$ is optimized for photo-consistency $\FF_\text{pc}$ between silhouette images and model density.

Figure \ref{fig:shapeReconstruction} shows the reconstructed shape after 100, 300 and 10000 gradient iterations from a 12th camera which is not used for optimization. This verifies that the geometry of an object can accurately be estimated in the same manner as the object pose. 
Note that a few Gaussians are pushed outside of the silhouette (presumably due to too large gradient steps) and vanish in size. These could be removed in a post-process.

The color of each Gaussian $G_q$ is inferred by the weighted average over the pixel colors of all pixels $(u, v)$ in all RGB input images, weighted by the corresponding Gaussian visibility $\V_q((u,v),\T)$.
This shape estimation of a human actor is a special instance of the actor model creation step proposed by Stoll \etal \cite{Stoll2011}, who optimize a parametric actor model that consists of Gaussians constrained to move with a skeleton. We show that our method is versatile enough to approximate the actor's shape without positional constraints between blobs.

\subsection{Additional tracking Experiment}
\label{sec:trackingQuality}

\begin{figure}%
\begin{center}
\includegraphics[width=0.8\linewidth]{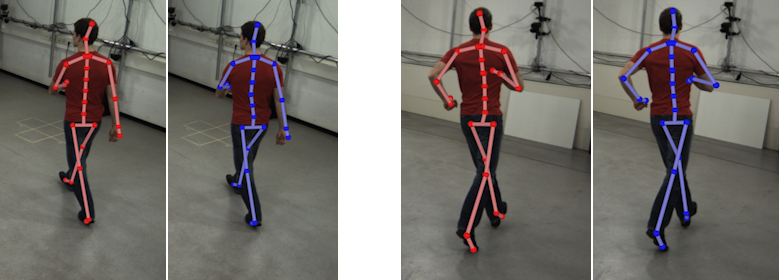}
\end{center}
\caption{Evaluation of our method on the \emph{Walker} sequence \cite{Stoll2011}. With only four cameras, our method (blue) is able to accurately track the whole sequence containing walking (left) and jogging (right) motions. Here compared to 12 camera tracking with the method of Stoll \etal (red).}
\label{fig:trackingQuality}
\end{figure}

We further evaluate the proposed visibility model on Stoll \etal's \emph{Walker} sequence \cite{Stoll2011}, for which no ground truth is available. 
Instead, a reimplementation of the method of Stoll \etal using 12 cameras is used as reference.
The same skeleton with 77 pose parameters and 43 Gaussians and image resolution 324$\times$243 is used for both methods.
The approach of Stoll \etal applied on the full available camera acquisition setup, consisting of 12 cameras, produces qualitatively comparable results to our approach applied on 4 cameras only,
see Figure \ref{fig:trackingQuality}. The average Euclidean 3D joint position distance between both results is only 2.55 cm. At the point of very fast arm-jogging motions with strong occlusions, small errors are visible, however, our method recovers quickly. Please watch the supplemental video for tracking results over the whole sequence.
 

\begin{figure}
\begin{tikzpicture}[tight background]
	\begin{axis}[footnotesize, width=\columnwidth, height=4.5cm,
		xmin=0, xmax=300, xlabel={Frame number}, xtick={50,100,...,400},
		ymin=0, ymax=170, ylabel={Euclidean distance [cm]}, ytick={0,20,...,200},
		legend columns=3, legend entries={ours $\FF_\text{mc}$, ours $\FF_\text{mc}'$ (quad-tree), Stoll \etal \cite{Stoll2011}},
		legend style={anchor=south,at={(0.45,1.08)},cells={anchor=west}},
		table/every table/.style={header=false,x expr=\coordindex}
	]%
\addplot[blue!70!black,  mark=none, semithick] table[y index=0] {images/data/ms_walk_2cam_resolution.txt};%
\addplot[green!70!black,   mark=none, semithick] table[y index=1] {images/data/ms_walk_2cam_resolution.txt};%
\addplot[red!70!black,   mark=none, semithick] table[y index=2] {images/data/ms_walk_2cam_resolution.txt};%

\end{axis}
\end{tikzpicture}
   \caption{Comparison of $\FF_\text{mc}$ and $\FF_\text{mc}'$, based on the Euclidean 3D joint position error of all limb joints with respect to marker based ground truth. The impact of using a hierarchical representation is much smaller than the improvement on Stoll \etal.}
\label{fig:imageResolutionFigure}
\end{figure}

\subsection{Input image resolution and thresholding}
\label{sec:inputImageResolutin}

The method of Stoll \etal operates on a hierarchical image representation, that clusters regions of similar color in a quad-tree, whereas our method operates on pixels.
To verify that the gained improvements are primarily due to our introduced visibility model, and not due to different image resolutions, we run our algorithm on the studio sequence
with the same quad-tree representation that models squared areas of similar color by a single pixel of corresponding size.
To make our energy model applicable to representations with varying pixel size, we construct the energy $\FF_\text{mc}'$, which is equivalent to $\FF_\text{mc}$, but weights each pixel by its area. 
The influence of the hierarchical representation on the reconstruction quality is much smaller than the improvement on Stoll \etal., as shown in Figure \ref{fig:imageResolutionFigure}.

Moreover, we validate that the error due to excluding model blobs with negligible contribution (Section 3.3 in the main paper) is vanishingly small; the average error across the first 100 frames of the \emph{Marker} sequence is increased by only 0.0038~cm (0.1\% of the total error).

\twocolumn[{
\section{Analytic gradients}
\label{sec:analyticGradients}

In this section, we provide the analytic gradients of the \emph{Gaussian visibility} $\V_q$ and the objective functions $\DD_\text{mc}$ and $\DD_\text{pc}$ with respect to the Gaussian parameters $\T$.
The gradients of $\DD_\text{mc}$ and $\DD_\text{pc}$ follow immediately from the visibility gradient:
\begin{equation}
\frac{\partial \DD_\text{mc}(\T,\I)}{\partial \gamma} \!=\!\! \sum_{(u,v)} \! \sum_q \! d(\I(u,v),a_q) \frac{\partial \V_q((u,v),\T)}{\partial \gamma} 
\text{,}
\label{eqn:dataTermBackprojection}
\end{equation}
and
\begin{align}
\frac{\partial \DD_\text{pc}(\T,\I)}{\partial \gamma} = 2 \sum_{(u.v) \in I} \sum_{i} \left(\hat{L}_i((u,v), \T) - \I_i(u,v) \right) \frac{\partial \hat{L}_i((u,v), \T)}{\partial \gamma},
\end{align}
where $i \! \in \! \{R,G,B\}$ ranges over the different color channels, and the radiance gradient is
\begin{align}
\frac{\partial \hat{L}(\cam,\n, \T)}{\partial \gamma}
= \sum_q \an_q \frac{\partial \V_q}{\partial \gamma} 
\text{.}
\end{align}

%
%
The gradient of visibility of Gaussian $G_q$ is 
\begin{align}
\frac{\partial \V_q(\cam,\n,\T)}{\partial \gamma} =  \sum_{s \in S_q} \frac{\partial \lambda_q}{\partial \gamma} T(\cam,\n,s,\T)  G_q(\cam+s \n)
+ \lambda_q \frac{\partial T(\cam,\n,s,\T)}{\partial \gamma} G_q(\cam+s \n)
+ \lambda_q T(\cam,\n,s,\T) \frac{\partial G_q(\cam+s \n)}{\partial \gamma} 
,
\end{align}
with $\lambda_q = \ell \sigma_q$, for some fixed $\ell$, it holds
\begin{align}
\frac{\partial \lambda_q}{\partial \sR_q} &= \ell, \frac{\partial \lambda_q}{\partial \mR_q} = 0.
\end{align}
The gradient of Gaussian density is
\begin{align}
\frac{\partial G_q(\cam+s \n)}{\partial \gamma} &= \frac{\partial \exp\!\left(-\frac{ ( s - \mR_q )^2}{2 \sR_q^2}\right) \cR_q}{\partial \gamma} 
.
\end{align}
In the following, we derive derivatives for all sub-terms with respect to the individual Gaussian parameters.
The sampling location $s$ depends on $\sR_q$ and $\mR_q$ of the Gaussian $G_q$ for which the visibility is inferred. In this case, $\frac{s-\mR_q}{\sR_q} = \frac{\mR_q + k \ell \sR_q - \mR_q}{\sR_q} = k \ell$ and
\begin{align}
\frac{\partial \cR_q  \exp\left(-\frac{(k\ell)^2}{2} \right) }{ \partial \mR_q} &= 0, \ \ \ 
\frac{\partial \cR_q  \exp\left(-\frac{(k\ell)^2}{2} \right) }{ \partial \cR_q} =\exp\left(-\frac{(k\ell)^2}{2} \right) , \ \text{ and }
 \frac{\partial \cR_q  \exp\left(-\frac{(k\ell)^2}{2} \right) }{ \partial \sR_q} =  0,
\end{align}
for parameters $\sR_p,\mR_p$ with $p \neq q$, the gradient $\frac{\partial G_q(\cam+s \n)}{\partial \gamma}$ is zero.
}] 
\twocolumn[{
The gradient of transmission $T$ is
\begin{align}
\frac{\partial T(\cam,\n,s,\T)}{\partial \gamma} &= T(\cam,\n,s,\T)
 \frac{\partial \sum_p \frac{\sR_p \cR_p}{\sqrt{\frac 2 \pi}} \left(\erf\!\left(\!\frac{-\mR_p}{\sqrt{2} \sR_p}\!\right) \!-\! \erf\!\left(\!\frac{s \!-\! \mR_p}{\sqrt{2} \sR_p}\!\right) \right)}{\partial \gamma}.
\label{eqn:visibilitySubTerms}
\end{align}
For the case of constant sampling location, i.e. $\frac{\partial s}{\partial \mR} = \frac{\partial s}{\partial \sR} = 0$, it holds 
\begin{align}
- \sqrt{\frac \pi 2} \frac{\partial  {\sR_p \cR_p \left(\erf\left(\frac{s-\mR_p}{\sqrt{2} \sR_p}\right) - \erf\!\left(\!\frac{-\! \mR_p}{\sqrt{2} \sR_p}\!\right)\right)} }{\partial \mR_p} 
&=  \cR_p \left(\exp\left(-\frac{(s-\mR_p)^2}{2 \sR_p^2} \right) - \exp\left(-\frac{(-\mR_p)^2}{2 \sR_p^2} \right) \right)
, \nonumber \\
- \sqrt{\frac \pi 2} \frac{\partial  {\sR_p \cR_p \left(\erf\left(\frac{s-\mR_p}{\sqrt{2} \sR_p}\right)- \erf\!\left(\!\frac{-\! \mR_p}{\sqrt{2} \sR_p}\!\right)\right)}}{\partial \cR_p} 
&= - \sqrt{\frac \pi 2} \sR_p  \left(\erf\left(\frac{s-\mR_p}{\sqrt{2} \sR_p}\right) - \erf\!\left(\!\frac{-\! \mR_p}{\sqrt{2} \sR_p}\!\right)\right), \nonumber \\
- \sqrt{\frac \pi 2} \frac{\partial  {\sR_p \cR_p \left(\erf\left(\frac{s-\mR_p}{\sqrt{2} \sR_p}\right)- \erf\!\left(\!\frac{-\! \mR_p}{\sqrt{2} \sR_p}\!\right)\right)}}{\partial \sR_p} 
&= - \sqrt{\frac \pi 2} \cR_p \left(\erf\left(\frac{s-\mR_p}{\sqrt{2} \sR_p}\right) - \erf\!\left(\!\frac{-\! \mR_p}{\sqrt{2} \sR_p}\!\right) \right) \nonumber \\
&+  \frac{\cR_p (s-\mR_p)}{\sR_p} \exp\left(-\frac{(s-\mR_p)^2}{2 \sR_p^2} \right)
+  \frac{\cR_p (-\mR_p)}{\sR_p} \exp\left(-\frac{(-\mR_p)^2}{2 \sR_p^2} \right).
\end{align}
And for the case of sampling locations $s$ depending on $G_q$, i.e. $\frac{\partial s}{\partial \mR} \neq 0$ and $\frac{\partial s}{\partial \sR} \neq 0$, we substitute  $\frac{s-\mR_q}{\sR_q} = \frac{\mR_q + k \ell \sR_q - \mR_q}{\sR_q} = k \ell$ as before and consider the special cases
%
%
%
\begin{align}
- \sqrt{\frac \pi 2} \frac{\partial  {\sR_q \cR_q \erf\left(\frac{k\ell}{\sqrt{2}}\right)}}{\partial \mR_q} 
&= 0
, \nonumber \\
- \sqrt{\frac \pi 2} \frac{\partial  {\sR_q \cR_q \erf\left(\frac{k\ell}{\sqrt{2}}\right)}}{\partial \cR_q} 
& = - \sqrt{\frac \pi 2} \sR_q \erf\left(\frac{s-\mR_q}{\sqrt{2} \sR_q}\right)
, \nonumber \\
- \sqrt{\frac \pi 2} \frac{\partial  {\sR_q \cR_q\erf\left(\frac{k\ell}{\sqrt{2}}\right)}}{\partial \sR_q} 
&= - \sqrt{\frac \pi 2} \cR_q \erf\left(\frac{s-\mR_q}{\sqrt{2} \sR_q}\right)
, \nonumber \\
-\sqrt{\frac \pi 2}      \sR_p \cR_p
\frac{\partial \erf\left(\frac{\mR_q + k\ell\sR_q -\mR_p}{\sqrt{2} \sR_p}\right)}{\partial \mR_q} 
&=  -\cR_q \exp\left(-\frac{(s -\mR_q)^2}{2 \sR_q}\right) \nonumber \\
-\sqrt{\frac \pi 2}      \sR_p \cR_p
\frac{\partial \erf\left(\frac{\mR_q + k\ell\sR_q -\mR_p}{\sqrt{2} \sR_p}\right)}{\partial \sR_q} 
&= - \cR_q k\ell \exp\left(-\frac{(s -\mR_q)^2}{2 \sR_q}\right).
\end{align}
Finally the derivatives of the ray density of Gaussian $G_q$ with respect to their 3D parameters are
\begin{align}
\frac{\partial \sR_q}{\partial \sigma_q} &= 1, \frac{\partial \sR_q}{\partial \mu_q} = 0, \nonumber \\
\frac{\partial \mR_q}{\partial \sigma_q}&= 0, \frac{\partial \mR_q}{\partial \mu_q}= n, \nonumber \\
\frac{\partial \cR_q}{\partial \sigma_q} &= \cR_q  \frac{(\cam-\m_q)^\top (\cam-\m_q) - \mR_q^2}{\sR_q^3} + \frac{\cR_q}{c_q} \frac{\partial c_q}{\partial \sigma_q}, \text{ and }\nonumber \\
\frac{\partial \cR_q}{\partial \m_q} &= \cR_q \frac{(\cam-\m_q) + \mR_q \n}{\sR_q^2}.
\end{align}
}] 
\clearpage


{\small
\bibliographystyle{ieee}
\bibliography{differentiableOcclusion}
}